%% file: Informative_Path_Planning_for_Parking_Occupancy_Estimation.tex
\documentclass[conference]{IEEEtran}
\IEEEoverridecommandlockouts
\usepackage{algorithmic}
\pdfoutput=1
\makeatletter

\newcommand{\Rmnum}[1]{\expandafter\@slowromancap\romannumeral #1@}
\makeatother
\renewcommand{\algorithmicrequire}{\textbf{Input:}}  
 
\input{configs}

\begin{document}

\title{Informative Path Planning of Autonomous Vehicle for Parking Occupancy Estimation}

\author{Yunze Hu$^{1\dag}$, Jiaao Chen$^{1\dag}$, Kangjie Zhou$^1$, Han Gao$^1$, Yutong Li$^2$, and Chang Liu$^{1*}$
\thanks{$\dag$Yunze Hu and Jiaao Chen contributed equally to this work.}
\thanks{$^1$Yunze Hu, Jiaao Chen, Kangjie Zhou, Han Gao and Chang Liu are with the Department of Advanced Manufacturing and Robotics, College of Engineering, Peking University, Beijing 100871, China (hu\_yun\_ze@stu.pku.edu.cn; jiaaochen@pku.edu.cn; kangjiezhou@pku.edu.cn; hangaocoe@pku.edu.cn; changliucoe@pku.edu.cn) (\textit{Corresponding author: Chang Liu.}).}
\thanks{$^2$Yutong Li is with the Department of Aerospace Engineering, University of Michigan, Ann Arbor, MI 48109, USA (wilson420813@gmail.com)}}

\maketitle
\thispagestyle{plain}
\pagestyle{plain}

\begin{abstract}
Parking occupancy estimation holds significant potential in facilitating parking resource management and mitigating traffic congestion. 
Existing approaches employ robotic systems to detect the occupancy status of individual parking spaces and primarily focus on enhancing detection accuracy through perception pipelines. 
However, these methods often overlook the crucial aspect of robot path planning, which can hinder the accurate estimation of the entire parking area.
In light of these limitations, we introduce the problem of informative path planning for parking occupancy estimation using autonomous vehicles and formulate it as a Partially Observable Markov Decision Process (POMDP) task. 
Then, we develop an occupancy state transition model and introduce a Bayes filter to estimate occupancy based on noisy sensor measurements. 
Subsequently, we propose the Monte Carlo Bayes Filter Tree, a computationally efficient algorithm that leverages progressive widening to generate informative paths. 
We demonstrate that the proposed approach outperforms the benchmark methods in diverse simulation environments, effectively striking a balance between optimality and computational efficiency.
\end{abstract}

\begin{IEEEkeywords}
Occupancy Estimation, Path Planning, Active Sensing, POMDP, Monte Carlo Tree Search.
\end{IEEEkeywords}

\section{Introduction}
The development of parking occupancy estimation techniques has emerged as a promising endeavor 
to enhance parking resource management and mitigate traffic congestion \cite{yang2017turning, varghese2019efficient,assemi2021street}.
A significant research focus has been directed towards the development of parking occupancy detection systems for pre-installed surveillance cameras, leveraging advancements in computer vision technologies\cite{nieto2018automatic, suhr2016automatic,9061155}. 
Nonetheless, the reliance on pre-installed cameras severely constrains the occupancy detection to parking spaces within the cameras' field of view, rendering it unable to provide accurate estimates for entire parking areas, let alone facilities without the presence of such cameras.

To overcome this limitation, recent studies start employing robotic systems, such as drones, to obtain accurate occupancy estimation on an ad hoc basis \cite{2020Quadrotor,8123131}. 
This approach is particularly suitable for parking areas that lack pre-installed sensors, such as street parking spaces and temporary parking areas for events.
For instance, Wang and Ren \cite{2020Quadrotor} developed an autonomous parking occupancy detection system that utilized a quadrotor to navigate the parking area and provide real-time occupancy data.
Zhou et al. \cite{8123131} proposed a vehicle counting method based on UAV images, which converts vehicle counting into estimating density values across image pixels.

While these methods offer increased mobility and flexibility for occupancy estimation, their primary focus lies in improving systems' perception pipeline to accurately identify the occupancy states of individual parking spaces \cite{nieto2018automatic, suhr2016automatic,90611552020Quadrotor,8123131}. 
However, the lack of consideration given to the path planning of robots in these methods can result in limited estimation accuracy of the entire parking area, since the path taken by the robot greatly influences the informativeness of sensor observations, which in turn affects the state estimation accuracy.
Furthermore, the utilization of drones requires dedicated labor and additional expenses for operational maintenance, further restricting the applicability of existing approaches.
 


To deal with these problems, this work focuses on informative path planning for an autonomous vehicle to accurately estimate the occupancy status of a parking area. 
The objective is to enable the vehicle to consistently and accurately estimate the occupancy states of all parking spaces by utilizing real-time measurements from the onboard sensor, and to dynamically plan an optimal path that maximizes the informativeness of sensor measurements, thereby enhancing the overall estimation accuracy across the entire parking area.
Importantly, as the population of intelligent vehicles with advanced sensing and planning capabilities continues to grow, utilizing such vehicles for occupancy estimation can present a commercially and technically feasible alternative to relying solely on drones.

State estimation is a vital component for accurately estimating parking occupancy, as it involves fusing real-time sensor measurements while considering the arrival and departure processes of vehicles at each parking space to determine the current occupancy status.
Filtering techniques, such as the Kalman filter, particle filter, and the more general Bayes filter framework \cite{thrun2005probabilistic},
are effective tools for state estimation in dynamic systems.
We propose modeling the state transition processes of parking spaces and estimating the parking occupancy status using the Bayes filter framework. This approach allows us to effectively capture the dynamics of parking space occupancy and provide reliable estimates.

Planning informative paths to enhance state estimation accuracy presents a general challenge \cite{Liu2017ModelPC,kantaros2021sampling, du2021parallelized, pinto2022multiagent}. 
One popular solution is formulating the path planning as a Partially Observable Markov Decision Process (POMDP). 
However, obtaining exact solutions to POMDPs is computationally intractable \cite{papadimitriou1987complexity}.
Therefore, many approximation algorithms have been proposed to obtain near-optimal solutions.
In recent years, sampling-based online algorithms utilizing Monte Carlo tree search (MCTS) \cite{Browne2012ASO} to solve POMDPs have achieved noteworthy success, particularly for large-scale problems. 
For example, Silver and Veness proposed POMCP that incorporates observation simulations and updates to MCTS during tree construction \cite{silver2010monte}. 
To tackle POMDPs with continuous or large action and observation spaces, Sunberg and Kochenderfer proposed POMCPOW and PFT-DPW, utilizing double progressive widening to restrict the number of child nodes, ensuring that simulations traverse the same node multiple times to prevent the search tree from being shallow \cite{Sunberg2017OnlineAF}. 
Based on PFT-DPW, Fischer and Tas developed the Information Particle Filter Tree that utilizes the information-theoretic reward function to generate informative paths for information gathering \cite{Fischer2020InformationPF} .
Despite the progresses made, these methods rely on specific approximation of belief states, such as particles, and cannot directly apply to informative path planning for parking occupancy estimation due to the substantial computational expenses associated with maintaining a large particle set to achieve satisfactory performance.

To address these limitations, this study proposes a systematic approach to informative path planning based on MCTS for an autonomous vehicle, aiming to accurately estimate the occupancy states of all parking spaces in a given parking area. 
The main contributions of the work are threefold: 
\begin{enumerate}
\item We define the problem of informative path planning for parking occupancy estimation and formulate it as a POMDP task. 
Additionally, We develop an occupancy state transition model and present a Bayes filter for occupancy estimation based on noisy sensor measurements.
\item We propose the Monte Carlo Bayes Filter Tree (MCBFT) algorithm for informative path planning. 
Specifically, we employ progressive widening in the observation space to avoid constructing a shallow policy tree. Furthermore, we design a heuristic policy
to guide the rollout procedure of MCTS, enabling the computationally efficient generation of informative paths.
\item Through extensive simulations conducted on various parking lot scenarios, we demonstrate that the solutions obtained by the MCBFT algorithm are consistent with the optimal solutions in over $90\%$ of scenarios. Moreover, the computational time of MCBFT is only around $30\%$ of the optimal algorithm, showcasing a favorable trade-off between optimality and computational efficiency.
\end{enumerate}


\section{Problem Formulation} 
Consider a two-dimensional parking lot with $N$ parking spaces. The autonomous vehicle is assumed to have access to the parking lot map, and is equipped with a sensor to perceive the occupancy status of the parking spaces within the field of view (FOV).
The autonomous vehicle is tasked with planning the optimal paths to observe and estimate the occupancy states of the whole parking area continually and accurately. 

\subsection{Environment Model}
\begin{figure}[!t]
\centering
\includegraphics[width=\linewidth]{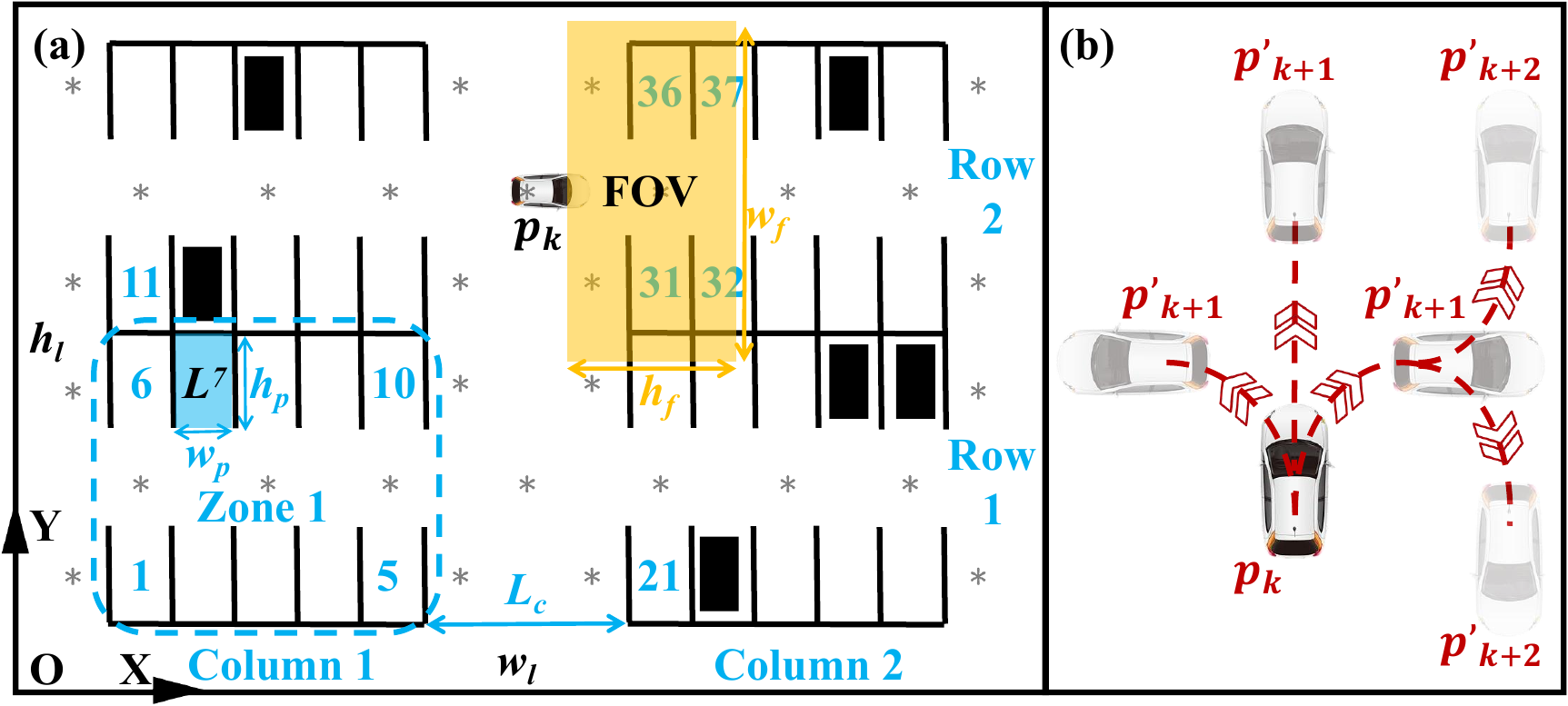}
\caption{\textbf{(a) Environment model and observation model.} The coordinate system is established with the origin $O$ and two coordinate axes $\overrightarrow{OX}$ and $\overrightarrow{OY}$. The parking lot is arranged in a $2$-by-$2$ zone configuration, with each zone containing $10$ parking spaces. The numbers $1$, $5$, $6$, $10$, $11$, $21$ represent the indices of parking spaces, and $L^7$ denotes the coverage set of the $7$th parking space. The vehicle is at pose $\boldsymbol{p_k}$ with the sensor FOV highlighted in orange, and $\mathcal{O}(\boldsymbol{p_k})=\{31,32,36,37\}$. The black solid rectangle represents that the corresponding parking space is ``occupied'', and the dark gray star symbols depict the discrete locations associated with the pose points in $V$.
\textbf{(b) Motion model.} The vehicle's current pose is $\boldsymbol{p_k}$ with three available actions: turning left, turning right, and moving forward. The potential future poses are depicted by vehicle icons with higher transparency.}
\label{fig:parkinglotmap}
\end{figure}

A two-dimensional Cartesian coordinate system is established in the parking lot (\cref{fig:parkinglotmap}(a)) that has a width of $w_l$ and a height of $h_l$, and is arranged in a grid of $r$ rows and $c$ columns, with a corridor of length $L_c$ located between each column. The parking lot is divided into $r\cdot c$ zones, each containing $n_z$ parking spaces. 
Each parking space is of rectangular shape, with the width denoted as $w_p$ and the height denoted as $h_p$.
The coordinates of the four vertices of the $i$th parking space are denoted as $(u^i_1, w^i_1)$, $(u^i_2, w^i_2)$, $(u^i_3, w^i_3)$, $(u^i_4, w^i_4)$, respectively, where $i\in\{1,\cdots,N\}$.
The 2D coverage set of the $i$th parking space is defined as $L^i \triangleq\{(u,w)\in\mathbb{R}^2\ |\ u=\sum_{j=1}^4\epsilon_ju^i_j,w=\sum_{j=1}^4\epsilon_jw^i_j,\sum_{j=1}^4\epsilon_j=1, \epsilon_j\in[0,1], j=1,\cdots,4\}\subset\mathbb{R}^2$. 
Let $\Delta t > 0$ represent the discretization time interval, and the $k$th time step corresponds to time $t_k$. 
The autonomous vehicle's pose at $k$th time step is denoted as $\boldsymbol{p}_k = (x_k, y_k, \theta_k)$, where $x_k$ and $y_k$ represent the position and $\theta_k$ denotes the orientation. 
A directed graph $G=(V, E)$ with $m$ vertices can be constructed such that $V = \{\boldsymbol{v}^1, \boldsymbol{v}^2,\ \cdots, \boldsymbol{v}^m\}$ represents the vehicle's discrete pose space, where $\boldsymbol{v}^j\in \mathbb{R}^3$ corresponds to the $j$th \textit{pose point} including the position and orientation with respect to the parking lot frame, and $E=V\times V$ denotes the directed edge set. If $\langle \boldsymbol{v}^i, \boldsymbol{v}^j\rangle \in E$, it indicates that the autonomous vehicle can travel from pose $\boldsymbol{v}^i$ to pose $\boldsymbol{v}^j$ within $\Delta t$. 


\subsection{Observation Model \label{subsec:Observation and Motion Model}}
Let $\boldsymbol{x}_k = (x^1_k, \cdots, x^N_k)\in \{0,1\}^{N}$ denote the ground-truth occupancy states, and $\boldsymbol{z}_k = (z^1_k, \cdots, z^N_k)\in \{0,1,\emptyset\}^{N}$ represent the sensor measurements of parking spaces. 
Here ``$1$'' means ``occupied'', ``$0$'' means ``unoccupied'', and ``$\emptyset$'' means the parking space is not within the sensor's FOV.
We adopt a rectangle-shaped area to model the sensor's FOV, noted as $\mathcal{F}(\boldsymbol{p}_k)\subset\mathbb{R}^2$.
Let $\mathcal{S}(\cdot)$ denote the area covered by a 2D set, and define $\mathcal{O}(\boldsymbol{p}_k)\subset \{1,\dots,N\}$ as the index set of parking spaces observable  by the vehicle at pose $\boldsymbol{p}_k$. The measurement $z_k^i$ of occupancy state $x^i_k$ is obtained when more than half of the parking space is within the sensor’s FOV, i.e.,
\begin{equation*}
    \mathcal{S} (L^i \cap \mathcal{F}(\boldsymbol{p}_k))\geq 0.5\mathcal{S}(L^i) \Leftrightarrow i\in \mathcal{O}(\boldsymbol{p}_k)\Leftrightarrow\;z_k^i\neq \emptyset.
\end{equation*}
The sensor follows a probabilistic observation model,
\begin{equation}
    p(z^i_k\ |\ x^i_k)=
    \begin{cases}
    1,& \text{ $ \text{if } z^i_k=\emptyset, i\notin \mathcal{O}(\boldsymbol{p}_k), $ }\\
    p_1,& \text{ $ \text{if } z^i_k=1, x^i_k = 1, i\in \mathcal{O}(\boldsymbol{p}_k), $ } \\
    1-p_1,& \text{ $ \text{if } z^i_k=0, x^i_k = 1, i\in \mathcal{O}(\boldsymbol{p}_k), $ }\\
    p_2,& \text{ $ \text{if } z^i_k=0, x^i_k = 0, i\in \mathcal{O}(\boldsymbol{p}_k), $ }\\
    1-p_2,& \text{ $ \text{if } z^i_k=1, x^i_k = 0, i\in \mathcal{O}(\boldsymbol{p}_k), $ }
    \end{cases}\label{eqn:subPerceptionModel}
\end{equation}
where parameters $p_1\in[0,1]$ and $p_2\in[0,1]$ represent the probability of correct observation of occupied and unoccupied parking spaces, respectively\footnote{Note that the actual observation model is sensor dependent and can be different from the one used in this work. However, the proposed method can be easily extended to sensors with different observation models.}.


\subsection{POMDP Model for Parking Occupancy Estimation}
The POMDP offers a powerful framework for sequential decision-making under uncertainty. 
A finite-horizon POMDP is defined by the 9-tuple $(D, S, A, \Omega, T, O, R, \gamma, b_0)$ \cite{Lauri2022PartiallyOM}, 
where the $S$, $A$, $\Omega$ are the state, action, and observation set, respectively.
Function $T$, $O$, $R$ represent the transition model, observation model, and reward function, respectively. Here $D$ is the planning horizon, $\gamma$ is the discount factor, and $b_0$ is the initial belief. We aim to find a policy $\boldsymbol{\pi^{*}}$ that can maximize the expected cumulative reward of the POMDP. 

Our work focuses on the sequential decision making for the autonomous vehicle with the objective of maximizing the acquisition of occupancy information within a predefined planning horizon, which can be modeled as a finite-horizon POMDP. 
The state and observation can be defined as $\boldsymbol{s}_k=(\boldsymbol{p}_k, \boldsymbol{x}_k)\in S$, $\boldsymbol{o}_k=\boldsymbol{z}_k\in\Omega$, respectively. Each pose $\boldsymbol{p}_k$ has a corresponding discrete action space $A(\boldsymbol{p}_k)$, in which each action $\boldsymbol{a}_k\in A(\boldsymbol{p}_k)$ drives the vehicle from pose $\boldsymbol{p}_k$ to another pose $\boldsymbol{p}_{k+1}\in\{\boldsymbol{p}\ |\ \langle \boldsymbol{p}_k,\boldsymbol{p}\rangle \in E\}$ following the vehicle motion model $\boldsymbol{p}_{k+1}=f(\boldsymbol{p}_{k},\boldsymbol{a}_k)$, as illustrated in \cref{fig:parkinglotmap}(b).
The transition model $T$ is defined as
\begin{equation}
\boldsymbol{s}_{k+1}=T(\boldsymbol{s}_k,\boldsymbol{a}_k)\label{eqn:TransitionModel},
\end{equation}
where $\boldsymbol{p}_{k+1}$ is propagated using the vehicle motion model $f$, and
the transition from $\boldsymbol{x}_k$ to $\boldsymbol{x}_{k+1}$ is based on the parking space state transition model $\sigma$, which will be specified in \Cref{subsec:occupancy_park_model}. 
The observation $\boldsymbol{o}_k=O(\boldsymbol{s}_k)$ 
is obtained based on \cref{eqn:subPerceptionModel}. 

Let $B_k = \{b^1_k, \cdots, b^N_k\}$ denote the vehicle's estimation of the occupancy state, with the belief state $b^i_k$ representing the probability of the $i$th parking space being occupied at time step $k$. 
An optimal policy sequence $\boldsymbol{\pi}^{*} = (\pi^{*}_k, \cdots, \pi^{*}_{k+D-1})$ is calculated so that the autonomous vehicle can plan an optimal path to efficiently estimate the parking occupancy, 
\begin{equation}
    \boldsymbol{\pi^{*}}= \mathop{\arg\max}\limits_{\boldsymbol{\pi}}\mathbb{E}\left[\sum_{t=k}^{k+D-1}\gamma^{t-k}R\ \bigg|\ B_k,\boldsymbol{p}_k,\boldsymbol{\pi}\right],\label{eqn:ObjectiveFunction}
\end{equation}
where $\mathbb{E}$ is the expectation operator, and the reward function $R$ is related with the estimation of the occupancy state, which will be specified in \Cref{MCBFT_reward}. The vehicle then executes the optimal action $\boldsymbol{a}_k^{*}=\pi^*_k(B_k,\boldsymbol{p}_k)\in A(\boldsymbol{p}_k)$ and repeats this process to replan at the next time step in the receding horizon manner, so that the vehicle can proactively generate informative paths based on the newly received observations.

\subsection{MCTS-Based Solutions to POMDPs}
MCTS provides a computationally efficient way to solve POMDPs \cite{Browne2012ASO}. MCTS incrementally constructs an asymmetric policy tree that comprises alternating layers of state nodes and action nodes through the repetition of four key phases: selection, expansion, simulation, and back-propagation. During the selection phase, a strategy that balances exploration and exploitation is employed to navigate through the current policy tree. Once a leaf node is reached, new nodes are added to the tree during the expansion phase. Then a simulation rollout is conducted from the leaf node to estimate the corresponding reward. Finally, in the back-propagation phase, the parameters of the visited nodes in this iteration are updated. Once the policy tree has been constructed, we can choose the optimal solution from the first layer of action nodes.

\section{Bayes Filtering for Occupancy Estimation}

\subsection{Parking Space State Transition Model}\label{subsec:occupancy_park_model}
We propose a discrete-time model for the transition of individual parking space states based on existing parking lot traffic models.
We assume the states of all parking spaces are independent of each other.

\subsubsection{Arrival}
The arrival of vehicles in a parking lot is usually modeled as a Poisson process \cite{Xiao2018HowLA}. In a similar vein, 
we propose to adopt a Poisson distribution to model the arrival process for each parking space.
Let $l$ denote the number of vehicles that pass the parking space during a time interval and are willing to park in the parking space if it is unoccupied. The Poisson distribution with mean parameter $\lambda$ can be expressed as:
\begin{equation}
P_a(l, \lambda)=\left\{
\begin{array}{rcl}
\frac{\lambda^{l}}{l!}e^{-\lambda} & & {\text{if}\ l = 0, 1, 2, ...} \\
0\ \ \   & & {\ \ \ \ \ \ \ \ \text{else}}
\end{array} \right..
\label{5}
\end{equation}

If a parking space is unoccupied at time $t_k$, then during time interval $[t_k, t_{k+1})$, the probability of no vehicle parking in the space is $P_a(l = 0, \lambda)$, and the transition probabilities can be defined as
\begin{subequations}
\begin{align}
p(x^i_{k+1}=0\ |\ x^i_{k}=0)=P_a(l=0, \lambda)=e^{-\lambda}
\label{eqn:ArrivalModel1},\\
p(x^i_{k+1}=1\ |\ x^i_{k}=0)=1-e^{-\lambda}.
\label{eqn:ArrivalModel2}
\end{align}
\label{eqn:ArrivalModel}
\end{subequations}
For narrative simplicity, we use $p_3$ to represent the transition probability \cref{eqn:ArrivalModel2}, i.e., $p_3\triangleq p(x^i_{k+1}=1\ |\ x^i_{k}=0)$.


\subsubsection{Departure}
For each parked vehicle, the parking time is usually modeled as an exponential distribution \cite{Xiao2018HowLA}. Let $\mu$ denote the departure rate for a parked vehicle, and the probability density function of the exponential distribution is
\begin{equation}
P_d(t)=\left\{
\begin{array}{rcl}
\mu e^{-\mu t} & & {\text{if}\ t > 0} \\
0\ \ \   & & {\ \ \text{else}}
\end{array} \right..
\label{eq:exp_pdf}
\end{equation}

In the case that a vehicle is parked in the parking space at time $t_k$, we can calculate the probability of the vehicle leaving the space during $[t_k, t_{k+1})$ through \cref{eq:exp_pdf}. The memoryless property of the exponential distribution leads to the following transition probabilities,
\begin{subequations}
\begin{align}
p(x^i_{k+1}=0\ |\ x^i_{k}=1)&=1-e^{-\mu \Delta t},
\label{eqn:DepartureModel1}\\
p(x^i_{k+1}=1\ |\ x^i_{k}=1)&=e^{-\mu \Delta t}.
\label{eqn:DepartureModel2}
\end{align}
\label{eqn:DepartureModel}
\end{subequations}
For narrative simplicity, we use $p_4$ to represent the transition probability \cref{eqn:DepartureModel2}, i.e., $p_4\triangleq p(x^i_{k+1}=1\ |\ x^i_{k}=1)$.

\cref{eqn:ArrivalModel,eqn:DepartureModel} comprise the parking space state transition model, which we compactly denote as
\begin{equation}
\label{eqn:transitionmodel2}
\boldsymbol{x}_{k+1}=\sigma(\boldsymbol{x}_k).
\end{equation}

\subsection{State Estimation with Bayes Filter}
We use the Bayes filter to recursively fuse sensor measurements to estimate the occupancy state of parking spaces.
The Bayes filter consists of two steps: prediction and update. 

\textbf{Prediction.} At time step $k$, each parking space's belief is forward predicted using the parking space state transition model \cref{eqn:transitionmodel2} in the prediction step,
\begin{equation}
\overline{b}^{i}_{k+1}=p_3(1-b^i_{k})+p_4b^i_{k}, i=1,\cdots,N.
\label{eqn:bayes_filter_prediction}
\end{equation}

\textbf{Update.} The posterior distribution of the occupancy state is calculated by fusing sensor measurements via the Bayes rule,
\begin{equation}
{b}^i_{k+1}=\eta p(z_{k+1}^i|x_{k+1}^i)\overline{b}^{i}_{k+1}, i=1,\cdots,N,
\label{eqn:bayes_filter_update}
\end{equation}
where $p(z_{k+1}^i|x_{k+1}^i)$ is the observation model defined in \cref{eqn:subPerceptionModel} and $\eta$ is the normalization factor.

\section{Monte Carlo Bayes Filter Tree for \\Path Planning}
\subsection{Reward Function Design}\label{MCBFT_reward}
A belief-based reward function $R$ is essential for quantifying the information gain associated with executing a specific action. 
In order to measure information uncertainty of parking occupancy status, we consider the entropy 
of the $i$th parking space, which can be written as
\begin{equation}
\mathcal{H}(b_k^{i}) = -b_k^{i}\mathrm{log_2}b_k^{i} - (1-b_k^{i})\mathrm{log_2}(1-b_k^{i}).
\label{eqn:Entropy}
\end{equation}
Then we select the negation of entropy $-\mathcal{H}$ as the information measure $\mathcal{I}$ and subsequently define the information gain as $\Delta \mathcal{I}(b,b') = \mathcal{I}(b') - \mathcal{I}(b)$, which has been widely used in the field of active sensing \cite{Mihaylova2002ACO}. Following this idea, the reward function is formulated as
\begin{equation}
    R=\Delta \mathcal{I}(B_k,B_{k+1}) = \sum_{i=1}^N(\mathcal{H}(b_k^{i}) - \mathcal{H}(b_{k+1}^{i})),\label{eqn:reward function}
\end{equation}
where $B_{k+1}$ is calculated through the Bayes filter \cref{eqn:bayes_filter_prediction,eqn:bayes_filter_update} from the occupancy estimation at the previous step $B_k$, action $\boldsymbol{a}_k=\pi_k(B_k,\boldsymbol{p}_k)$ and observation $\boldsymbol{z}_{k+1}$.

\subsection{Traversal Path Planning}

\begin{algorithm}[!t]  
        \caption{Traversal Algorithm}  
        \begin{algorithmic}[1] 
        \small
            \REQUIRE $B_k:$ the set of beliefs, $\Psi:$ the set of all possible paths
            \ENSURE $\psi^{*}:$ the path selected
                \FOR{$j=1:N_T$}
                \STATE $\Delta H_j=0$
                \FOR {$i=1:N$}
                \STATE $N_0^{i,1} \gets (1,b^{i}_k)$
                \STATE $H_j^{i,0}=\mathcal{H}(b_k^i)$
                \FOR{$d= 1:D$}
                \FOR{$e=1:m_{d-1}$}
                \STATE $({\phi}_{d-1}^{i,e},b_{k+d-1}^{i,e})\gets N_{d-1}^{i,e}$
                \STATE $\overline{b}_{k+d}^{i,e}=p_3(1-b_{k+d-1}^{i,e})+p_4 b_{k+d-1}^{i,e}$
                \IF{$i\in \mathcal{O}(\boldsymbol{p}_{k+d}^{j})$}
                \STATE $b_{k+d}^{i,e}=\eta p_1\overline{b}_{k+d}^{i,e}$
                \STATE 
                $\xi=p_1\overline{b}_{k+d}^{i,e}+(1-p_2)(1-\overline{b}_{k+d}^{i,e})$
                
                \STATE $N_{d}^i \gets N_{d}^i\cup\{({\phi}_{d-1}^{i,e}\xi,b_{k+d}^{i,e}),({\phi}_{d-1}^{i,e}(1-\xi),1-b_{k+d}^{i,e})\}$
                
                \ELSE
                \STATE$N_{d}^i \gets N_{d}^i\cup({\phi}_{d-1}^{i,e},\overline{b}_{k+d}^{i,e})$
                \ENDIF
                \ENDFOR
                \ENDFOR
                \STATE $\Delta H_j^i=0$
                \FOR{$d= 1:D$}
                \STATE $H_j^{i,d}=0$
                \FOR{$e=1:m_d$}
                \STATE $H_j^{i,d}=H_j^{i,d}+{\phi}_d^{i,e}\mathcal{H}(b_{k+d}^{i,e})$
                \ENDFOR
                \STATE $\Delta H_j^i=\Delta H_j^i+\gamma^{d-1}(H_j^{i,d-1}-H_j^{i,d})$
                \ENDFOR
                \STATE $\Delta H_j=\Delta H_j+\Delta H_j^i$
                \ENDFOR
                \ENDFOR
                \STATE $j^{*}=\mathop{\arg\max}\limits_{j}\Delta H_{j}$
                \STATE $\psi^*=\psi_{j^*}$
        \end{algorithmic}
        \label{alg:traverse_algo}
\end{algorithm}

A straightforward strategy for path planning is to enumerate all feasible paths and compare their associated cumulative rewards. The path with the highest cumulative reward, as defined in \cref{eqn:ObjectiveFunction}, is selected.
This Traversal path planning algorithm is presented in \cref{alg:traverse_algo}.


First we generate all $N_T$ feasible paths for the upcoming $D$ planning steps according to the current pose $\boldsymbol{p}_{k}$ and the environment model. Denote $\Psi=\{\psi_1,\cdots,\psi_{N_T}\}$ as the set of all the feasible paths, and each path $\psi_j=\{\boldsymbol{p}_{k}^j, \boldsymbol{p}_{k+1}^j,\cdots,\boldsymbol{p}_{k+D}^j\}$ consists of the pose sequence of the vehicle within the planning horizon $D$.
For the $i$th parking space, a search tree for simulating possible future observations is produced. The nodes in the $d$th-layer of the tree are described as $N_d^i=\{N_d^{i,1},\cdots, N_d^{i,m_d}\}$, where the $e$th node $N_d^{i,e} = ({\phi}_d^{i,e},b_{k+d}^{i,e})$ consists of the probability ${\phi}_d^{i,e}$ of the realized observation sequence from planning step $1$ to $d$ and the corresponding belief $b_{k+d}^{i,e}$, and $m_d$ is the total number of distinct observation sequences between step $1$ and $d$. Note that the growth of new nodes in $N_d^i$ from $N_{d-1}^i$ is determined by the observation of the $i$th parking space at planning step $d$. 

For each path, the search tree for the $i$th parking space is built recursively based on Bayes filter considering all possible future observations. We first set the initial node $N_0^{i,1}=(1,b_k^{i})$ (Line 4). At planning step $d$, we extract the information $(\phi_{d-1}^{i,e},b_{d+k-1}^{i,e})$ from the $e$th node $N_{d-1}^{i,e}$ that has been calculated in the $(d-1)$th layer (Line 8) , and use them to compute the belief $\overline{b}_{d+k}^{i,e}$ based on the prediction step \cref{eqn:bayes_filter_prediction} (Line 9). If the vehicle can observe the parking space, the sensor observation model is utilized in the update step to calculate the belief $b_{d+k}^{i,e}$ according to \cref{eqn:bayes_filter_update} (Line 11). Furthermore, we add two new nodes to consider the two possible observation results at planning step $d$ (Line 12-13), 
where $\xi$ denotes the probability of the ``occupied'' observation. If the vehicle cannot observe the parking space, only one node is created at planning step $d$, and the update step of the Bayes filter is skipped (Line 15). 
Once all the nodes are expanded to the $D$th layer, we compute the cumulative entropy reduction $\Delta H^i_j$ of the $i$th parking space for the whole prediction horizon (Line 19-26). 
Then the entropy reduction of all parking spaces are subsequently summed to compute the overall entropy reduction of the current path $\Delta H_j$ (Line 27). 
Finally, we select the optimal path $\psi^*=\{\boldsymbol{p}_{k}^*, \boldsymbol{p}_{k+1}^*, \cdots, \boldsymbol{p}_{k+D}^*\}$ with the highest cumulative entropy reduction of the parking lot $\Delta H_{j^*}$ (Line 30-31) and choose the associated pose $\boldsymbol{p}^*=\boldsymbol{p}_{k+1}^*$ as the next pose point for the vehicle to head for.

\subsection{Monte Carlo Bayes Filter Tree for Path Planning} 
While the Traversal algorithm can obtain optimal solutions through enumeration, the time complexity exponentially increases with growing planning horizon $D$. 
Inspired by \cite{Sunberg2017OnlineAF, Fischer2020InformationPF}, we propose MCBFT, an MCTS-based algorithm for informative path planning, which enables us to evaluate parking occupancy under low computational burden with minor accuracy loss compared to the Traversal algorithm, as shown in \cref{alg:Information Bayesian Filter Tree}.
The \textsc{Search} procedure serves as the entry point for the decision-making process. It takes the initial belief $B_k$ and the vehicle's initial pose $\boldsymbol{p}_k$ as inputs, constructs a policy tree, and selects the optimal action $\boldsymbol{a}^{*}$ when the simulation times reach the preset number $I$.

\begin{algorithm}[!t]
    \caption{Monte Carlo Bayes Filter Tree} 
    \label{alg:Information Bayesian Filter Tree}
    \begin{algorithmic}[1] 
        \small
        \renewcommand{\algorithmicrequire}{ \textbf{Procedure}}
        \REQUIRE \textsc{Search}$(B_k,\boldsymbol{p}_k)$
                \FOR {$i\in1:I$}
                \STATE \textsc{Simulate}($B_k, \boldsymbol{p}_k, 0$)
                \ENDFOR
                \RETURN $\boldsymbol{a}^{*}=\mathop{\arg\max}\limits_{\boldsymbol{a}'}Q(B_k,\boldsymbol{p}_k, \boldsymbol{a}')$
                
        \renewcommand{\algorithmicrequire}{ \textbf{Procedure}}
        \REQUIRE \textsc{Simulate}$(B, \boldsymbol{p}, h)$
            \IF{$h=D$} 
                \RETURN $0$
            \ELSE
            \IF{$|C(B,\boldsymbol{p})|<|A(\boldsymbol{p})|$} 
                    \STATE $\boldsymbol{a} \gets$ sample from $A(\boldsymbol{p})-C(B,\boldsymbol{p})$
                    \STATE $C(B,\boldsymbol{p}) \gets C(B,\boldsymbol{p})\cup\{\boldsymbol{a}\}$
                \ENDIF 
                \STATE $\boldsymbol{a}^*=\mathop{\arg\max}\limits_{\boldsymbol{a}'}Q(B,\boldsymbol{p},\boldsymbol{a}')+c\sqrt{\frac{\log q(B,\boldsymbol{p})}{q(\boldsymbol{a}')}}$\label{alg:argmax}
            \IF{$|C(\boldsymbol{a}^*)|<\kappa q(\boldsymbol{a}^{*})^{\delta}$}  
                \STATE $\boldsymbol{x} \gets$ sample from $B$\\
                \STATE $(\boldsymbol{p}',\boldsymbol{x}') \gets T((\boldsymbol{p},\boldsymbol{x}),\boldsymbol{a}^*)$\\
                \STATE $\boldsymbol{o} \gets O(\boldsymbol{p}',\boldsymbol{x}')$\\
                \STATE $B' \gets$ \textsc{Bayes Filter}$(B,\boldsymbol{o})$\\
                \STATE $\omega(B',\boldsymbol{p}') \gets p(\boldsymbol{o}\ |\ B,\boldsymbol{p}')$\\
                \IF{$(B',\boldsymbol{p}') \notin C(\boldsymbol{a}^*)$}  
                    \STATE $C(\boldsymbol{a}^*) \gets C(\boldsymbol{a}^*)\cup \{(B',\boldsymbol{p}')\}$
                    \STATE $J \gets \Delta I(B,B')+\gamma$\textsc{Rollout}($B', \boldsymbol{p}', h+1$)
                \ELSE 
                    \STATE $J \gets \Delta I(B,B')+\gamma$\textsc{Simulate}($B', \boldsymbol{p}', h+1$)
                \ENDIF
            \ELSE  
                \STATE $(B',\boldsymbol{p}') \gets$ sample $(B',\boldsymbol{p}')$ w.p. $\frac{\omega(B',\boldsymbol{p}')}{\sum_{\overline{B'}}{\omega(\overline{B'},\boldsymbol{p}')}}$
                \STATE $J \gets \Delta I(B,B')+\gamma$\textsc{Simulate}($B', \boldsymbol{p}', h+1$)
            \ENDIF 
            \STATE $q(B,\boldsymbol{p}) \gets q(B,\boldsymbol{p})+1$
            \STATE $q(\boldsymbol{a}^*) \gets q(\boldsymbol{a}^*)+1$
            \STATE $Q(B,\boldsymbol{p},\boldsymbol{a}^*) \gets Q(B,\boldsymbol{p},\boldsymbol{a}^*)+\frac{J-Q(B,\boldsymbol{p},\boldsymbol{a}^*)}{q(\boldsymbol{a}^*)}$ 
        \ENDIF
        \RETURN $J$

        \renewcommand{\algorithmicrequire}{ \textbf{Procedure}}
        \REQUIRE \textsc{Rollout}$(B, \boldsymbol{p}, h)$
            \IF{$h=D$} 
                \RETURN $0$
            \ELSE
                \STATE $\boldsymbol{a}^*\gets$ \textsc{RolloutPolicy}($B,\boldsymbol{p}$)\\
                \STATE $\boldsymbol{x} \gets$ sample from $B$\\
                \STATE $(\boldsymbol{p}',\boldsymbol{x}') \gets T((\boldsymbol{p},\boldsymbol{x}),\boldsymbol{a}^*)$\\
                \STATE $\boldsymbol{o} \gets O(\boldsymbol{p}',\boldsymbol{x}')$\\
                \STATE $B' \gets$ \textsc{Bayes Filter}$(B,\boldsymbol{o})$\\
                \STATE $J \gets \Delta I(B,B')+\gamma$\textsc{Rollout}($B', \boldsymbol{p}', h+1$)
            \ENDIF
            \RETURN $J$
    \end{algorithmic}  
\end{algorithm}  

The policy tree comprises two distinct types of nodes: belief nodes and action nodes. The former contains a tuple of the occupancy estimation and the vehicle's pose, denoted as $(B,\boldsymbol{p})$, while the latter consists of the vehicle's action $\boldsymbol{a}$. Compared to PFT-DPW \cite{Sunberg2017OnlineAF} and IPFT \cite{Fischer2020InformationPF} that utilize the particle filter to approximate the belief, MCBFT stores precise belief in each belief node $(B,\boldsymbol{p})$, which allows for the exact calculation of the total belief-based reward $J$. We maintain a record $\langle q,C,\boldsymbol{o},\omega \rangle$ for each belief node $(B,\boldsymbol{p})$ in the tree, which consists of its number of visits $q$, set of child nodes $C$, corresponding observation $\boldsymbol{o}$, and probability $\omega$ of generating $\boldsymbol{o}$ from the observation space. Likewise, for each action node $\boldsymbol{a}$, we track $\langle q,C,Q \rangle$, which includes its number of visits $q$, set of child nodes $C$, and its estimated reward $Q$ based on belief $B$ and pose $\boldsymbol{p}$, where $B$ and $\boldsymbol{p}$ are obtained from the parent node of action node $\boldsymbol{a}$. 

The \textsc{Simulate} function progressively constructs the search tree of depth $h\in\{0, 1, \cdots, D\}$. An action node $\boldsymbol{a}^*$ is selected according to the upper confidence bound (UCB) (Line 12) \cite{Browne2012ASO}. 
As the number of parking spaces that the autonomous vehicle can observe increases, the observation space increases exponentially, and the probability of sampling the same observation decreases exponentially, leading to constructing a shallow tree. 
To prevent this issue, we use progressive widening to limit the number of belief nodes linked to the same parent node $\boldsymbol{a}^*$ to $\kappa q(\boldsymbol{a}^*)^{\delta}$, where $\kappa$ and $\delta$ are preset hyper-parameters (Line 13), so that the simulations can pass through the same child node multiple times. If the action node $\boldsymbol{a}^*$ does not have enough child nodes, a new observation $\boldsymbol{o}$ is generated using the observation model $O$, and the posterior belief $B'$ is computed with the Bayes filter (Line 16-17). The $\omega$ is calculated based on the observation model \cref{eqn:subPerceptionModel} (Line 18). On the other hand, if the action node $\boldsymbol{a}^*$ already has enough child nodes, an existing belief child node is sampled proportional to $\omega$ (Line 26). If the node $(B',\boldsymbol{p}')$ has already been created, the \textsc{Simulate} function is called recursively to visit and create deeper nodes (Line 23, 27). Otherwise, the \textsc{Rollout} function is performed to estimate $J$ (Line 21, 34-44). 
We introduce a heuristic method in \textsc{Rollout} to efficiently generate informative paths. 
Specifically in \textsc{RolloutPolicy} function (Line 37), instead of random action selection, the optimal action is chosen probabilistically based on the Traversal algorithm with a preset planning horizon $D'$.
Finally, the parameters of visited nodes are updated (Line 29-31).

\section{Simulation Results and Analyses}
\label{simulation}
The proposed path planning approach is evaluated on various parking lot scenarios. 
First, three manually designed, representative scenarios are created to serve as a sanity check, qualitatively validating the proposed approach in generating paths that align with intuitive expectations.
Then, $200$ random scenarios are generated to quantitatively evaluate the generality of the proposed algorithm.

\subsection{Simulation Setup}
We utilize the Matlab Automated Parking Valet toolbox to construct parking lots. We construct three distinct parking lot models, the parameters of which are listed in \cref{tab:parameter of models}. 
The locations of pose points are distributed in the corridors, inside the zones, and around the periphery of the zones, and then the pose points are interconnected based on the motion model of the vehicle.
Each parking space measures $w_p=3m$ and $h_p=6m$, and the parking space state transition model \cref{eqn:transitionmodel2} takes parameters $\mu=0.000378$ and $\lambda=0.000624$, which is derived from the actual parking lot data \cite{Xiao2018HowLA}. 
The dimensions of FOV are $w_f=20m$ and $h_f=10m$, and the observation model takes parameters $p_1=p_2=0.95$.
The parameters in MCBFT are set as $I=100$ and $D'=5$. 
The simulations are conducted on a laptop with Intel Core i7-1065G7 CPU@1.30GHz and 16GB RAM.

\subsection{Qualitative Evaluation Using Three Representative Scenarios}
\begin{figure}[!t]
\centering
\includegraphics[width=\linewidth]{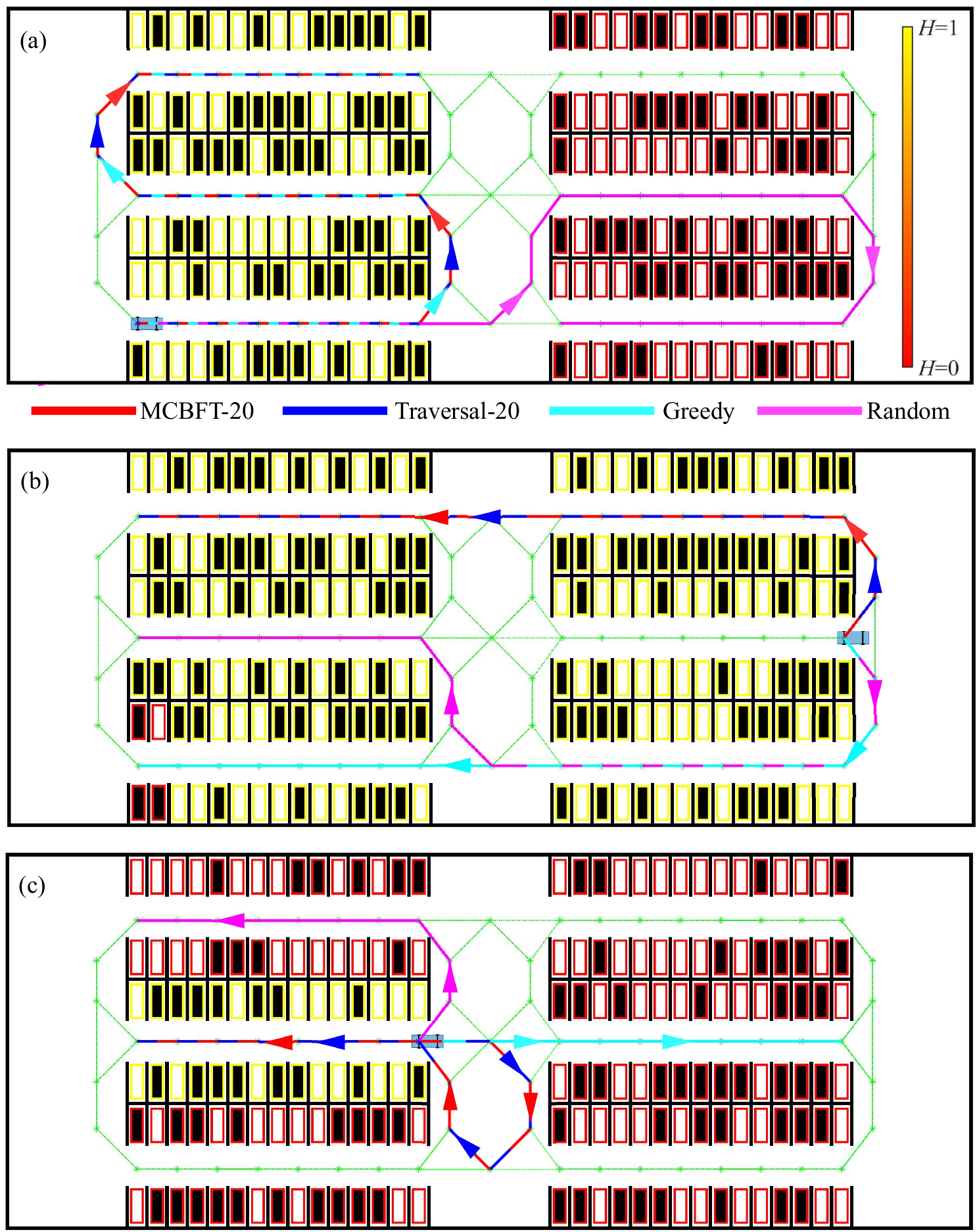}
\caption{\textbf{Path comparison in three representative scenarios.} The red, blue, cyan, and purple lines represent paths planned by MCBFT-$20$, Traversal-$20$, the Greedy algorithm and the Random Walk algorithm, respectively. The colored rectangle outlining each parking space represents the initial entropy, which ranges from $0$ to $1$ as illustrated in the colorbar in (a). (a) Scenario \Rmnum{1}. (b) Scenario \Rmnum{2}. (c) Scenario \Rmnum{3}.}\label{Qualitative Evaluation}
\end{figure}

We design three representative scenarios based on Parking Lot Model \Rmnum{1} to evaluate the ability of the proposed algorithm in planning informative paths. 
For each scenario, we conduct $10$ simulations, where the parking space state transitions and the sensor measurements are simulated stochastically. 
We denote the $30$ parking spaces in the bottom left, bottom right, mid left, mid right, top left and top right area of the parking lot as the zone \Rmnum{1} to zone \Rmnum{6}, respectively. 
If the $i$th parking space is not observed a priori, then $b_{0}^i$ is set to be $0.5$. Otherwise, if the $i$th parking space is initially observed as occupied or unoccupied, then $b_{0}^i$ is initialized as $0.95$ or $0.05$, respectively.
We evaluate the performance of the MCBFT and the Traversal algorithm, both with a planning horizon of $20$ steps, namely MCBFT-$20$ and Traversal-$20$, respectively. 
The baseline algorithms include the Greedy algorithm, which essentially corresponds to the Traversal algorithm with planning horizon $D = 1$, and the Random Walk algorithm, where the vehicle randomly selects an action at each pose. \cref{Qualitative Evaluation} visualizes the three scenarios and the generated paths under different path planning approaches. 

\begin{table}[!t]
\caption{Parking Lot Model Parameters}
\label{tab:parameter of models}
\begin{center}
\resizebox{0.7\linewidth}{!}{
\begin{tabular}{|c|c|c|c|c|c|c|}
\hline
\textbf{Model}& $w_l(m)$& $h_l(m)$& $L_c(m)$ & $r$ &$c$ & $n_z$\\
\hline
\textbf{\textit{Model \Rmnum{1}}}& $144$& $56$& $18$ & $3$ & $2$ & $30$ \\
\hline
\textbf{\textit{Model \Rmnum{2}}}& $108$& $130$& $18$ & $7$ & $2$ & $18$ \\
\hline
\textbf{\textit{Model \Rmnum{3}}}& $153$& $74.5$& $18$ & $4$ & $3$ & $18$ \\
\hline
\end{tabular}}
\label{tab1}
\end{center}
\end{table}

\subsubsection{Scenario \Rmnum{1}}
The initial pose of the autonomous vehicle is set as $(19.5,9.0,0.0)$, and  
the right half of the parking lot is observed a priori (\cref{Qualitative Evaluation}(a)).  
The paths generated by MCBFT-$20$, Traversal-$20$, and the Greedy algorithm 
can all accomplish sequential observations of the unobserved zones, thus rapidly decreasing the information uncertainty. 
In contrast, in $90\%$ of the simulations, the Random Walk algorithm produces unfavorable paths that repeatedly detect zones that have already been observed, leading to small reduction in entropy.

\subsubsection{Scenario \Rmnum{2}}
The autonomous vehicle's pose is now initialized at $(124.5,28.0,0.0)$, and only the four parking spaces in zone \Rmnum{1} are previously observed (\cref{Qualitative Evaluation}(b)).
To maximize the decrease of uncertainty, it is more reasonable for the autonomous vehicle to make a left turn and observe zone \Rmnum{6} and zone \Rmnum{5} sequentially, which corresponds to the blue Traversal planning path shown in \cref{Qualitative Evaluation}(b). 
In $8$ out of the $10$ simulation outcomes, MCBFT-$20$ generates the same path as Traversal-$20$. 
In contrast, when utilizing the Greedy algorithm or the Random Walk algorithm, only in $4$ out of $10$ simulations does the vehicle first observe zone \Rmnum{6} and \Rmnum{5}.

\subsubsection{Scenario \Rmnum{3}}
The autonomous vehicle starts at pose $(61.5,28.0,0.0)$, and is most uncertain about the parking occupancy status of zone \Rmnum{3} (\cref{Qualitative Evaluation}(c)). 
Under this circumstance, both Traversal-$20$ and MCBFT-$20$ choose to turn around and observe zone \Rmnum{3} first.
In contrast, in the Greedy and the Random Walk algorithm, the number of simulations where the vehicle performs such behavior is only $0$ and $1$ out of $10$, respectively, and in the remaining simulations, the vehicle tends to stay in the already observed zones, resulting in little information gain.

 
\subsection{Quantitative Evaluation via Randomly Generated Scenarios}
To further evaluate the generality of the proposed approach, we consider two different parking lot models, Parking Lot Model \Rmnum{2} and \Rmnum{3}, and randomly generate $100$ scenarios for each model. 
For each scenario, we randomly select a set of zones to be pre-observed. 
Observed zones are initially assigned random belief values between $0.3$ and $0.95$ for ``occupied'' spaces, and between $0.05$ and $0.7$ for ``unoccupied'' spaces. 
Beliefs of unobserved zones are set to be $0.5$. 
The pose of the autonomous vehicle is also randomly initialized. 

We evaluate the performance of five algorithms: the MCBFT with a horizon of $10$ steps, namely MCBFT-$10$, the Greedy algorithm, the Random Walk algorithm, and the Traversal algorithm with planning horizons of 10 and 5 steps, namely Traversal-$10$ and Traversal-$5$, respectively.
All tested algorithms share the same ground-truth occupancy status time series $(\boldsymbol{x}_0,\cdots,\boldsymbol{x}_K)$, where $K$ denotes the simulation length and is set to be three-fourths of the total number of discrete locations reachable by the vehicle. 
We record the entropy $H_k$ of the belief and the correctness rate $\alpha_k$ of the estimation, defined as the ratio of the parking spaces where the estimation is consistent with the actual state. 
Specifically, the vehicle estimates spaces with a belief exceeding $0.6$ as ``occupied'', and spaces with a belief below $0.4$ as ``unoccupied''. When the belief lies between $0.4$ and $0.6$, the vehicle cannot confidently estimate the state of parking space, resulting in an ``unsure'' estimation that is inconsistent with any actual state.

After simulation, we evaluate the performance of the aforementioned five algorithms based on the three metrics: the improvement of the correctness rate $\Delta \alpha \triangleq \alpha_K - \alpha_0$, the average computational time $\overline{\tau}$ ($\text{s}/ \text{step}$), and the entropy reduction percentage $\frac{\Delta H}{H_0}$, where $\Delta H$ is defined as $H_0 - H_K$ that represents the decrease in entropy.
Specifically, we denote 
$N_{\alpha}, N_{\Delta H}$ as the number of scenarios in which MCBFT-$10$ outperforms another algorithm in terms of $\Delta \alpha$ and $\frac{\Delta H}{H_0}$, respectively. 
To further compare the path decisions of MCBFT and the Traversal algorithm, we define the pose points where multiple actions are available as the \textit{decision points}, and introduce the concept of consistency rate $\beta$, which is defined as the ratio of decision points where the MCBFT-$10$ and Traversal-$10$ take the same action. 

\begin{figure}[!t]
\centering
\includegraphics[width=\linewidth]{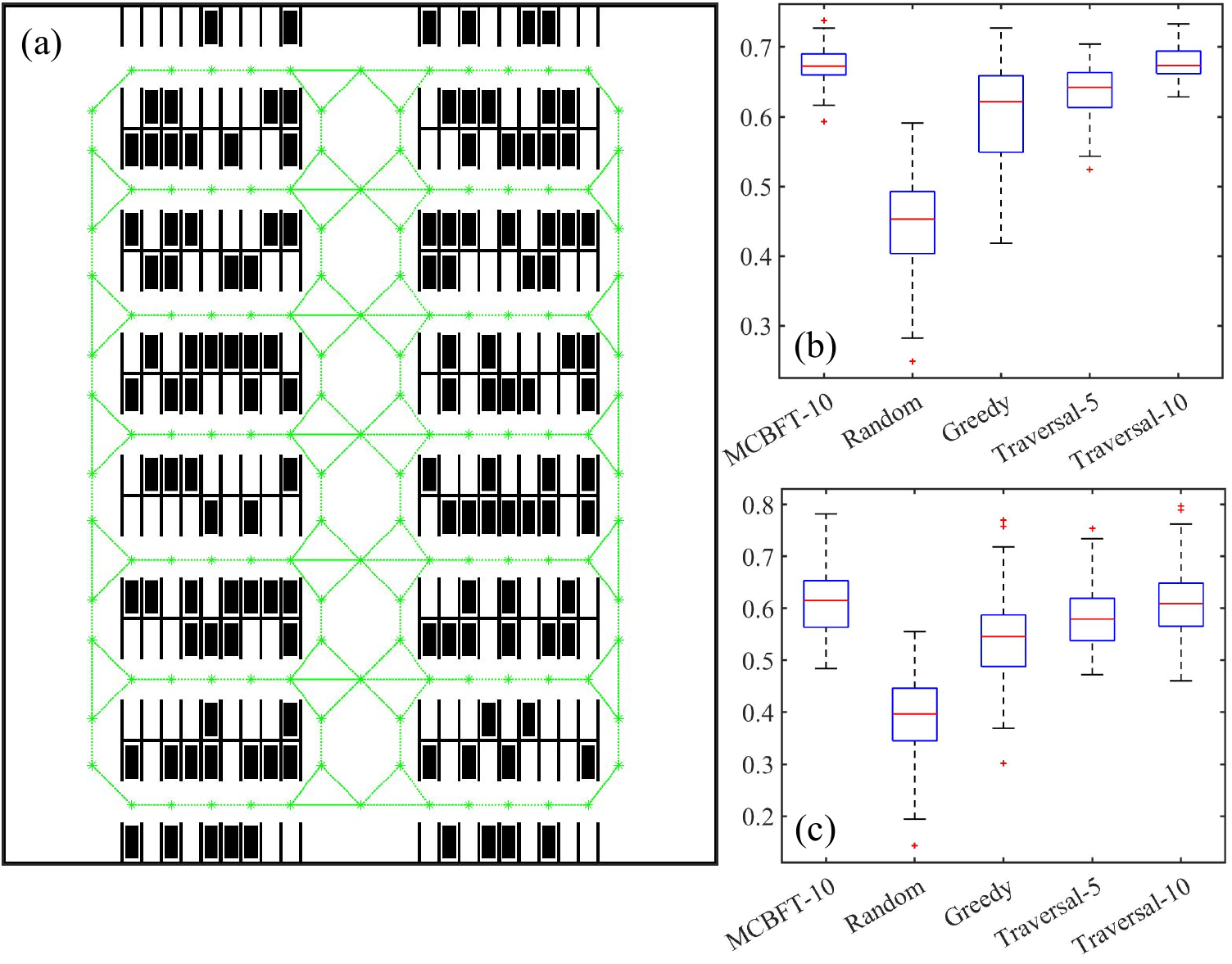}
\caption{(a) Illustration of Parking Lot Model \Rmnum{2}. (b) and (c) compare the simulation results of five algorithms in terms of entropy reduction percentage $\frac{\Delta H}{H_0}$ and improvement of correctness rate $\Delta\alpha$, respectively.}\label{tall model}
\end{figure}

\begin{table}[!t]
\caption{Simulation Results in Parking Lot Model \Rmnum{2}}
\label{tab:tall model}
\begin{center}
\resizebox{0.95\linewidth}{!}{
\begin{tabular}{|c|c|c|c|c|c|}
\hline
\textbf{Alg}& \textbf{\textit{MCBFT-10}}& \textbf{\textit{Traversal-10}}& \textbf{\textit{Traversal-5}}& \textbf{\textit{Greedy}}& \textbf{\textit{Random Walk}} \\
\hline
$\overline{\tau}$& $0.35$& $0.97$& $4.8\times10^{-2}$& $2.7\times10^{-3}$& $1.6\times10^{-5}$ \\
\hline
$N_{\alpha}$& $/$& $45$& $73$& $84$& $100$ \\
\hline
$N_{\Delta H}$& $/$& $45$& $79$& $81$& $100$ \\
\hline
\end{tabular}}
\label{tab1}
\end{center}
\end{table}

\subsubsection{Parking Lot Model \Rmnum{2}}\label{subsubsec:sim_tall_park_lot}
We consider a parking lot model (\cref{tall model}(a)) that has four more rows than Parking Lot Model \Rmnum{1}.
The number of locations accessible to the vehicle is $125$, and the simulation runs for $94$ time steps.
As \cref{tall model} shows, MCBFT-$10$ and Traversal-$10$ outperform the other algorithms on both metrics: $\Delta \alpha$ and $\frac{\Delta H}{H_0}$. 
As shown in \cref{tab:tall model}, MCBFT-$10$ demonstrates superior performance compared to Traversal-$5$ and the Greedy algorithm in most of the scenarios. In nearly half of the scenarios, the performance of MCBFT-$10$ surpasses that of Traversal-$10$. Moreover, MCBFT-$10$ demonstrates an average running time that is approximately $36\%$ of that required by Traversal-$10$.
Out of the $2634$ decision points, MCBFT-$10$ and Traversal-$10$ make the same decisions in $2465$ instances, resulting in $\beta=93.58\%$. 
These results demonstrate that MCBFT can obtain optimal actions in the majority of cases while achieving high computational efficiency. 


\subsubsection{Parking Lot Model \Rmnum{3}}
\begin{figure}[!t]
\centering
\includegraphics[width=\linewidth]{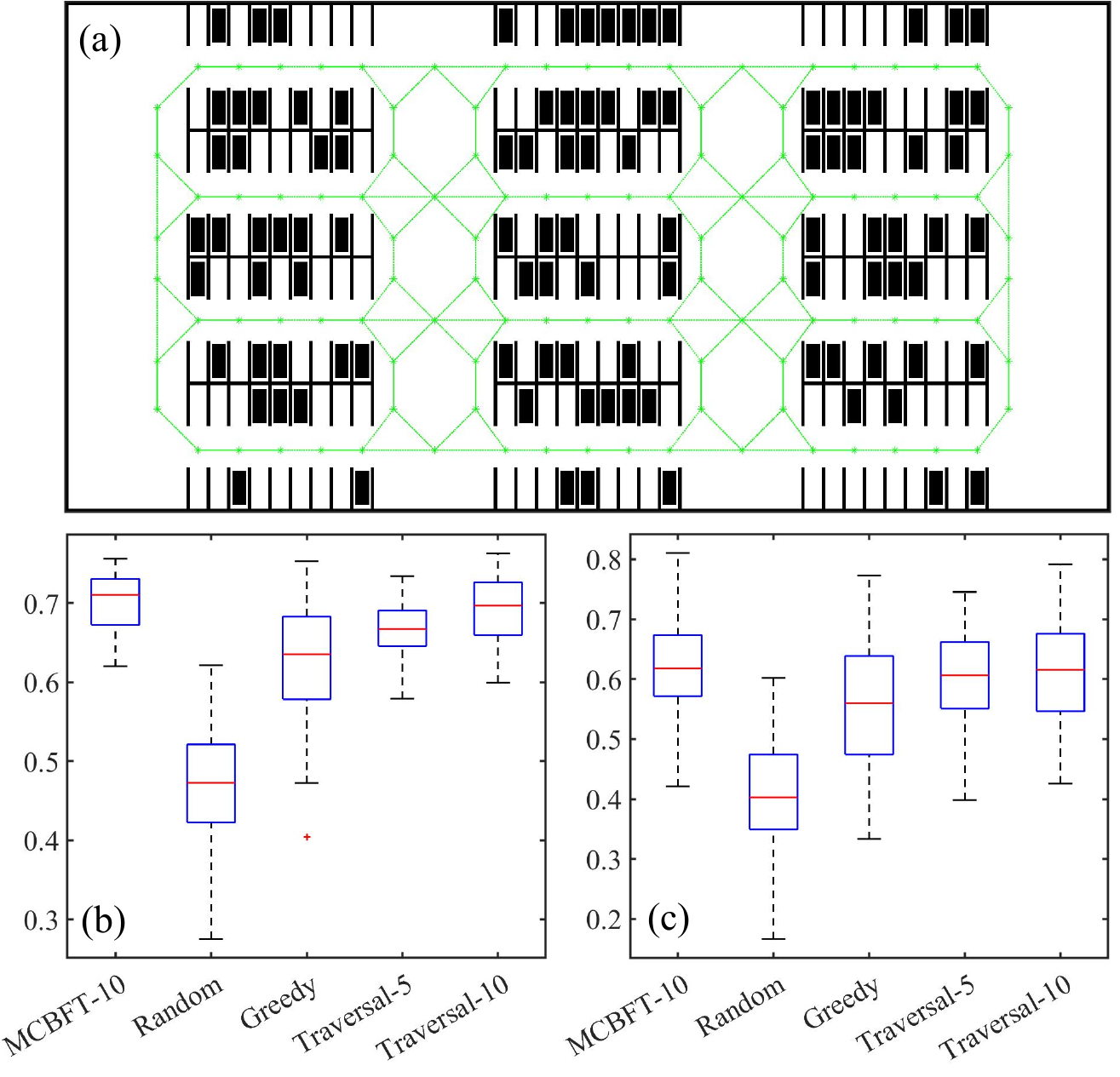}
\caption{(a) Illustration of Parking Lot Model \Rmnum{3}. (b) and (c) compare the simulation results of five algorithms in terms of entropy reduction percentage $\frac{\Delta H}{H_0}$ and improvement of correctness rate $\Delta\alpha$, respectively.}\label{flat model}
\end{figure}

We design a parking lot model (\cref{flat model}(a)) that has one more column and row than the Parking Lot Model \Rmnum{1}.
The number of locations the vehicle can travel to is $104$, and the simulation runs for $78$ time steps. The results, as depicted in \cref{flat model} and \cref{tab:flat model}, show similar trends as \Cref{subsubsec:sim_tall_park_lot}. 
The $\beta$ reaches $92.84\%$, as MCBFT-$10$ and Traversal-$10$ select the same option in $2413$ out of $2599$ decision points.


\begin{table}[!t]
\caption{Simulation Results in Parking Lot Model \Rmnum{3}}
\label{tab:flat model}
\begin{center}
\resizebox{0.95\linewidth}{!}{
\begin{tabular}{|c|c|c|c|c|c|}
\hline
\textbf{Alg}& \textbf{\textit{MCBFT-10}}& \textbf{\textit{Traversal-10}}& \textbf{\textit{Traversal-5}}& \textbf{\textit{Greedy}}& \textbf{\textit{Random Walk}} \\
\hline
$\overline{\tau}$& $0.20$& $0.67$& $4.0\times10^{-2}$& $2.3\times10^{-3}$& $1.2\times10^{-5}$ \\
\hline
$N_{\alpha}$& $/$& $55$& $75$& $84$& $100$ \\
\hline
$N_{\Delta H}$& $/$& $53$& $77$& $82$& $100$ \\
\hline
\end{tabular}}
\label{tab1}
\end{center}
\end{table}




\section{Conclusion}
We present the problem of informative path planning for autonomous vehicles to estimate parking occupancy and formulate it under the POMDP framework. 
We develop MCBFT that utilizes progressive widening to mitigate the high computational cost and enable online path planning. 
Simulation results show that MCBFT achieves a favorable trade-off between optimality and computational efficiency.
Future work includes evaluating the proposed approach in realistic parking lot environments and extension to multi-vehicle parking occupancy estimation.

{
\small
\bibliography{ref}
\bibliographystyle{ieeetr}
}

\end{document}

%% file: configs.tex
\usepackage{graphicx} \graphicspath{ {figures/} }
\usepackage{amsmath,amssymb,mathabx,amsbsy,mathtools,etoolbox}
\usepackage[utf8]{inputenc} 
\usepackage[T1]{fontenc}    

\usepackage{algorithmic}
\usepackage[ruled,linesnumbered]{algorithm2e}
\usepackage{acronym}
\usepackage{enumitem}
\usepackage[breaklinks,colorlinks,linkcolor=black]{hyperref}
\usepackage{balance}
\usepackage{xspace}
\usepackage{setspace}
\usepackage[skip=3pt,font=small]{subcaption}
\usepackage[skip=3pt,font=small]{caption}
\usepackage[dvipsnames,svgnames,x11names]{xcolor}
\usepackage[capitalise,nameinlink]{cleveref}
\usepackage{booktabs,tabularx,colortbl,multirow,multicol,array,makecell}
\usepackage{pifont}
\usepackage{cite}
\usepackage{nicematrix}

\makeatletter
\DeclareRobustCommand\onedot{\futurelet\@let@token\@onedot}
\def\@onedot{\ifx\@let@token.\else.\null\fi\xspace}

\makeatother

\makeatletter
\def\BState{\State\hskip-\ALG@thistlm}
\makeatother

\makeatletter
\renewcommand{\paragraph}{%
  \@startsection{paragraph}{4}%
  {\z@}{0ex \@plus 0ex \@minus 0ex}{-1em}%
  {\hskip\parindent\normalfont\normalsize\bfseries}%
}
\makeatother

\crefname{algorithm}{Alg.}{Algs.}
\Crefname{algocf}{Algorithm}{Algorithms}
\crefname{section}{Sec.}{Secs.}
\Crefname{section}{Section}{Sections}
\crefname{table}{Tab.}{Tabs.}
\Crefname{table}{Table}{Tables}
\crefname{figure}{Fig.}{Fig.}
\Crefname{figure}{Figure}{Figure}

\definecolor{gblue}{HTML}{4285F4}
\definecolor{gred}{HTML}{DB4437}
\definecolor{ggreen}{HTML}{0F9D58}

\definecolor{mygray}{gray}{.92}

\acrodef{qp}[QP]{Quadratic Programming}
\acrodef{fd}[FD]{Force Decomposition}
\acrodef{ros}[ROS]{Robot Operating System}
\acrodef{uav}[UAV]{Unmanned Aerial Vehicle}
\acrodef{dof}[DoF]{Degree-of-freedom}
\acrodef{com}[CoM]{Center-of-Mass}
\acrodef{ilc}[ILC]{Iterative Learning Control}
\acrodef{siso}[SISO]{single-input-single-output} 
\acrodef{zpetc}[ZPETC]{zero-phase-error tracking controller}
\acrodef{mimo}[MIMO]{multi-input-multi-output}
\def\proof{\@IEEElegacywarn{proof}{IEEEproof}\IEEEproof}

\let\oldnl\nl
\newcommand{\nonl}{\renewcommand{\nl}{\let\nl\oldnl}}%
\makeatother

%% file: Informative_Path_Planning_for_Parking_Occupancy_Estimation.bbl
\begin{thebibliography}{10}

\bibitem{yang2017turning}
S.~Yang and Z.~S. Qian, ``Turning meter transactions data into occupancy and
  payment behavioral information for on-street parking,'' {\em Transportation
  Research Part C: Emerging Technologies}, vol.~78, pp.~165--182, 2017.

\bibitem{varghese2019efficient}
A.~Varghese and G.~Sreelekha, ``An efficient algorithm for detection of vacant
  spaces in delimited and non-delimited parking lots,'' {\em IEEE Transactions
  on Intelligent Transportation Systems}, vol.~21, no.~10, pp.~4052--4062,
  2019.

\bibitem{assemi2021street}
B.~Assemi, A.~Paz, and D.~Baker, ``On-street parking occupancy inference based
  on payment transactions,'' {\em IEEE Transactions on Intelligent
  Transportation Systems}, vol.~23, no.~8, pp.~10680--10691, 2021.

\bibitem{nieto2018automatic}
R.~M. Nieto, A.~Garcia-Martin, A.~G. Hauptmann, and J.~M. Martinez, ``Automatic
  vacant parking places management system using multicamera vehicle
  detection,'' {\em IEEE Transactions on Intelligent Transportation Systems},
  vol.~20, no.~3, pp.~1069--1080, 2018.

\bibitem{suhr2016automatic}
J.~K. Suhr and H.~G. Jung, ``Automatic parking space detection and tracking for
  underground and indoor environments,'' {\em IEEE Transactions on Industrial
  Electronics}, vol.~63, no.~9, pp.~5687--5698, 2016.

\bibitem{9061155}
R.~Ke, Y.~Zhuang, Z.~Pu, and Y.~Wang, ``A smart, efficient, and reliable
  parking surveillance system with edge artificial intelligence on iot
  devices,'' {\em IEEE Transactions on Intelligent Transportation Systems},
  vol.~22, no.~8, pp.~4962--4974, 2021.

\bibitem{2020Quadrotor}
Y.~Wang and B.~Ren, ``Quadrotor-enabled autonomous parking occupancy
  detection,'' in {\em IEEE/RSJ International Conference on Intelligent Robots
  and Systems (IROS)}, 2020.

\bibitem{8123131}
H.~Zhou, L.~Wei, M.~Fielding, D.~Creighton, S.~Deshpande, and S.~Nahavandi,
  ``Car park occupancy analysis using uav images,'' in {\em 2017 IEEE
  International Conference on Systems, Man, and Cybernetics (SMC)},
  pp.~3261--3265, 2017.

\bibitem{thrun2005probabilistic}
S.~Thrun, W.~Burgard, and D.~Fox, {\em Probabilistic Robotics}.
\newblock MIT Press, 2005.

\bibitem{Liu2017ModelPC}
C.~Liu and J.~K. Hedrick, ``Model predictive control-based target search and
  tracking using autonomous mobile robot with limited sensing domain,'' {\em
  2017 American Control Conference (ACC)}, pp.~2937--2942, 2017.

\bibitem{kantaros2021sampling}
Y.~Kantaros, B.~Schlotfeldt, N.~Atanasov, and G.~J. Pappas, ``Sampling-based
  planning for non-myopic multi-robot information gathering,'' {\em Autonomous
  Robots}, vol.~45, no.~7, pp.~1029--1046, 2021.

\bibitem{du2021parallelized}
B.~Du, K.~Qian, C.~Claudel, and D.~Sun, ``Parallelized active information
  gathering using multisensor network for environment monitoring,'' {\em IEEE
  Transactions on Control Systems Technology}, vol.~30, no.~2, pp.~625--638,
  2021.

\bibitem{pinto2022multiagent}
S.~C. Pinto, S.~B. Andersson, J.~M. Hendrickx, and C.~G. Cassandras,
  ``Multiagent persistent monitoring of targets with uncertain states,'' {\em
  IEEE Transactions on Automatic Control}, vol.~67, no.~8, pp.~3997--4012,
  2022.

\bibitem{papadimitriou1987complexity}
C.~H. Papadimitriou and J.~N. Tsitsiklis, ``The complexity of markov decision
  processes,'' {\em Mathematics of Operations Research}, vol.~12, no.~3,
  pp.~441--450, 1987.

\bibitem{Browne2012ASO}
C.~Browne, E.~J. Powley, D.~Whitehouse, S.~M.~M. Lucas, P.~I. Cowling,
  P.~Rohlfshagen, S.~Tavener, D.~P. Liebana, S.~Samothrakis, and S.~Colton, ``A
  survey of monte carlo tree search methods,'' {\em IEEE Transactions on
  Computational Intelligence and AI in Games}, vol.~4, pp.~1--43, 2012.

\bibitem{silver2010monte}
D.~Silver and J.~Veness, ``Monte-carlo planning in large pomdps,'' {\em
  Advances in Neural Information Processing Systems}, vol.~23, 2010.

\bibitem{Sunberg2017OnlineAF}
Z.~Sunberg and M.~J. Kochenderfer, ``Online algorithms for pomdps with
  continuous state, action, and observation spaces,'' in {\em International
  Conference on Automated Planning and Scheduling}, 2017.

\bibitem{Fischer2020InformationPF}
J.~Fischer and {\"O}.~S. Tas, ``Information particle filter tree: An online
  algorithm for pomdps with belief-based rewards on continuous domains,'' in
  {\em International Conference on Machine Learning}, 2020.

\bibitem{Lauri2022PartiallyOM}
M.~Lauri, D.~Hsu, and J.~Pajarinen, ``Partially observable markov decision
  processes in robotics: A survey,'' {\em IEEE Transactions on Robotics},
  vol.~39, pp.~21--40, 2022.

\bibitem{Xiao2018HowLA}
J.~Xiao, Y.~Lou, and J.~Frisby, ``How likely am i to find parking? – a
  practical model-based framework for predicting parking availability,'' {\em
  Transportation Research Part B: Methodological}, 2018.

\bibitem{Mihaylova2002ACO}
L.~S. Mihaylova, T.~Lefebvre, H.~Bruyninckx, K.~Gadeyne, and J.~D. Schutter,
  ``A comparison of decision making criteria and optimization methods for
  active robotic sensing,'' in {\em Numerical Methods and Application}, 2002.

\end{thebibliography}
